\documentclass[11pt,a4paper]{article}
\usepackage[hyperref]{acl2017}
\usepackage{times}
\usepackage{latexsym}
\usepackage{times}
\usepackage{epsfig}
\usepackage{graphicx}
\usepackage{amsmath}
\usepackage{amssymb}
\usepackage{enumitem}
\usepackage{caption}
\usepackage{multirow}
\usepackage{url}

\aclfinalcopy 

\title{Linguistic Markers of Influence in Informal Interactions \vspace{2ex}}

\author{ Shrimai Prabhumoye\thanks{\quad Both authors contributed equally to this work.} \textsuperscript{1}, \quad
  Samridhi Choudhary \footnotemark[1] \textsuperscript{1}, \quad
 Evangelia Spiliopoulou \textsuperscript{1} \vspace{-3.5ex} \\
  \AND
 Christopher Bogart \textsuperscript{2}, \quad  
 Carolyn Penstein Rose \textsuperscript{1}, \quad  
 Alan W Black \textsuperscript{1} \\
 {\tt \{sprabhum,sschoudh,espiliop,cbogart,cprose,awb\}@andrew.cmu.edu} \vspace{-2ex} \\
 \\\AND
 \textsuperscript{1} \textnormal{Language Technologies Institute} \\
  Carnegie Mellon University \\
    \\\And
  \textsuperscript{2} \textnormal{Institute for Software Research} \\
  Carnegie Mellon University \\
 }

\begin{document}
\maketitle
\begin{abstract}
There has been a long standing interest in understanding `Social Influence' both in Social Sciences and in Computational Linguistics. In this paper, we present a novel approach to study and measure interpersonal influence in daily interactions. Motivated by the basic principles of influence, we attempt to identify indicative linguistic features of the posts in an online knitting community. We present the scheme used to operationalize and label the posts with indicator features. Experiments with the identified features show an improvement in the classification accuracy of influence by 3.15\%. Our results illustrate the important correlation between the characteristics of the language and its potential to influence others.
\end{abstract}

\section{Introduction}

Influence is a topic of great interest in the Social Sciences. Social Influence is defined as a situation where a person's thoughts, feelings or behaviors are affected by the real or imagined presence of others \cite{cialdini2004social}. In their study of social influence research, \citet{cialdini2002science} identify six basic principles that govern how one person might influence another. They are: liking, reciprocation, consistency, scarcity, social validation and authority. These principles control how influence plays out in different social situations.

The above mentioned principles constitute a solid basis for most of the work in this domain. Prior computational approaches for understanding influence, have primarily focused on influence as an explicit intention of the people involved \cite{tan2016winning,biran2012detecting,simfriends}. In this paper, we study influence from a different perspective: influence in daily, interpersonal interactions. We explore different language features based on the aforementioned theoretical principles and their correlation with influence. We attempt to extend the prior computational efforts on social influence, by using insights from the Social Sciences.

Influence can be defined and operationalized in different settings. A majority of computational work on interpersonal influence focuses on the analysis of social networks that employ probabilistic methods to analyze and maximize the flow of influence in these networks. There have been recent efforts in understanding influence in social media conversations with the aim of finding influential people \cite{biran2012detecting, quercia2011mood,rosenthal2016social}. We investigate what we can learn from language about influence from informal interactions where there is no explicit motivation to influence others. We look at user interactions in a social networking website for people interested in knitting, weaving, crocheting and fiber arts called Ravelry \footnote{https://www.ravelry.com/}, which is a large DIY online community with tens of thousands of sub-communities within it.

In the following sections we talk about prior work on social influence and the approaches taken to study it. We describe our dataset and the task setup that allows us to measure influence. We give an overview of the linguistic features we identified, inspired from theoretical insights of social influence. Finally, we present our results and conclude with discussion.

\section{Related Work}
There has been a substantial amount of computational work on modeling and detecting influence that can be broadly divided in two categories: `\textit{Influence in Social Networks}' and `\textit{Influence in Interactions}', each of which we discuss in this section. The aforementioned six principles play a pivotal role in defining relevant tasks for modeling and detecting influence. An example research question is: `Do people, who are connected in a social network and who like each other, display social influence (`liking principle') through their (correlated) activities in the network?' \cite{anagnostopoulos2008influence}. 

\subsection{Influence in Social Networks}
The computational models of influence in social networks primarily focus on ‘influence quantification’ and ‘influence diffusion’. \citet{goyal2010learning} present different probabilistic models (static, dynamic and discrete-time models) for quantifying influence between users in Flickr. They study how people are influenced by the actions of others, especially their social contacts, when performing actions (like joining a community). Their work quantifies the interplay of the principles of ‘Social Validation’ and ‘Liking’ and its effects on the decisions made by community members. \citet{tang2009social} use a Topical Affinity Propagation (TAP) model to quantify topic based social influence in large networks. The model is based on the the idea that the users in a social network are influenced by others for different reasons. They attempt to differentiate social influences from different angles (topics). \citet{anagnostopoulos2008influence} design time shuffling experiments to verify the existence of social influence as a driving factor behind activities observed in social networks.

Twitter has been a favorite target for such network analyses too \cite{weng2010twitterrank, shuai2012modeling, bakshy2011everyone, cha2010measuring}. For example, \citet{anger2011measuring} measure influence on Twitter as the social network potential of users. They  look for different influence indicators, like compliance, identification, internalization and neglect.

Traditional communication theory \citep{rogers2010diffusion} has stated that a small group of individuals, called `influentials', have better skills and excel at persuading others. Therefore, targeting these influential individuals in a network can be expected to result in a widespread chain reaction of influence with small cost \citep{katz1966personal}. The computational efforts based on this theory attempts to find a subset of nodes in a network (aka seed nodes) that would maximize the diffusion or the spread of influence. \citet{chen2009efficient} explore different algorithms and heuristics to maximize influence in a network. \citet{goyal2011data} introduce the ‘credit distribution’ model, which uses a data based approach to maximize influence by looking at historical data.

These efforts to model probability and diffusion of influence primarily focus on task-level actions relevant to the social network and not on the content of interaction between the participants. The following subsection details prior work on modeling influence based on the content of conversations.

\subsection{Influence in Interactions}
Bales and colleagues \cite{bales1956task, bales197315}, developed the idea that language is a form of contribution to group interaction that functions as a resource for maintaining group cohesion. In this direction, \citet{reid2000conversation} study conversations in small groups in order to investigate how conversational turns can be used to exert influence. Their analysis supports the idea that perceived influence is positively correlated with speakers' number of utterances \cite{ng1993gaining} and their successful interruptions \cite{ng1995interruption}. They modeled influential language as language that is aligned to the norms and the goals of the group; in other words was `prototypical' to the group. Their study found that speakers who use utterances and interruptions with high content prototypicality achieve a higher influence ranking. 

Other efforts use linguistic style choices and dialog patterns to detect influence in a conversation \cite{simfriends, quercia2011mood, nguyen2014modeling, rosenthal2016social}. They study influence and influential language through dialog structure, sentiment, valence, persuasion, agreement and control of conversational topics on online corpora. For example, \citet{biran2012detecting} explore communication characteristics that make someone an opinion leader or influential in online conversations. They model influential language by studying the conversational behaviors. They find that specific patterns in dialog like: initiating new topics of conversation, contributing more to dialog than others and engendering longer dialog threads on the same topic, are associated with higher influence. 

Language has also been explored as a resource for other tasks. \citet{tan2016winning}, for example, explore how different language factors may indicate persuasiveness in an online community (ChangeMyView) on Reddit. They study the effect of stylistic choices in the presentation of an argument that can make it more persuasive. 

As mentioned earlier, the majority of these approaches view influence as an important motivation behind the conversations. Our work attempts to study interpersonal influence as it occurs naturally among peers, without an explicit motivation to influence others. We explore the effect of language on influence, based on the theoretical principles.

\section{Data}

Our analysis is based on the posts written by the users of an online knitting platform called Ravelry. It is a social networking website for people interested in knitting, weaving, crocheting, spinning and more. It is ideal for large-scale data analysis as it has more than 6 million members, with 50,000 users being added every month. It provides a rich platform for textual analysis of social interactions, as it is a host to roughly billions of posts, thousands of user groups and discussion forums from different parts of the world. 

This is a community of people who have a shared interest in fiber arts. Members use this platform to create groups and forums. Some of these groups target people with specific characteristics, for example: groups for beginners, groups for people with heart conditions, groups for men who like to knit.  Members discuss and share their ideas, projects and collections of yarn, fiber and things that they find interesting. People generally borrow knitting patterns from other members and adapt them for their own projects. Therefore, the social dynamics of this community affords people the opportunity to share their interests and learn from each other.

These features of the community make the platform suitable for studying social influence in interpersonal interactions. We can observe the language used in a post, the members exposed to it, and the number of members who use a project pattern (which we refer to as a knitting pattern) mentioned in the post for their own project. These form the foundation for the approach described below.

\subsection{Operationalization of Influence}
Ravelry allows us to maintain information about the knitting pattern used in a project and the time stamps of the posts in a thread. Using this information, we can identify the knitting pattern adopted by a user and the posts that mention the pattern. This helps us to link a post and the knitting pattern mentioned in it to the users who adopted and potentially adapted the pattern after it was posted. We study these posts in order to identify the indicative linguistic features that lead to the pattern uptake. Therefore, we operationalize both the `users exposed' and the `pattern uptake'.


\subsubsection{Exposure}

\begin{figure}[h]
\centering
	\includegraphics[width=0.45\textwidth]{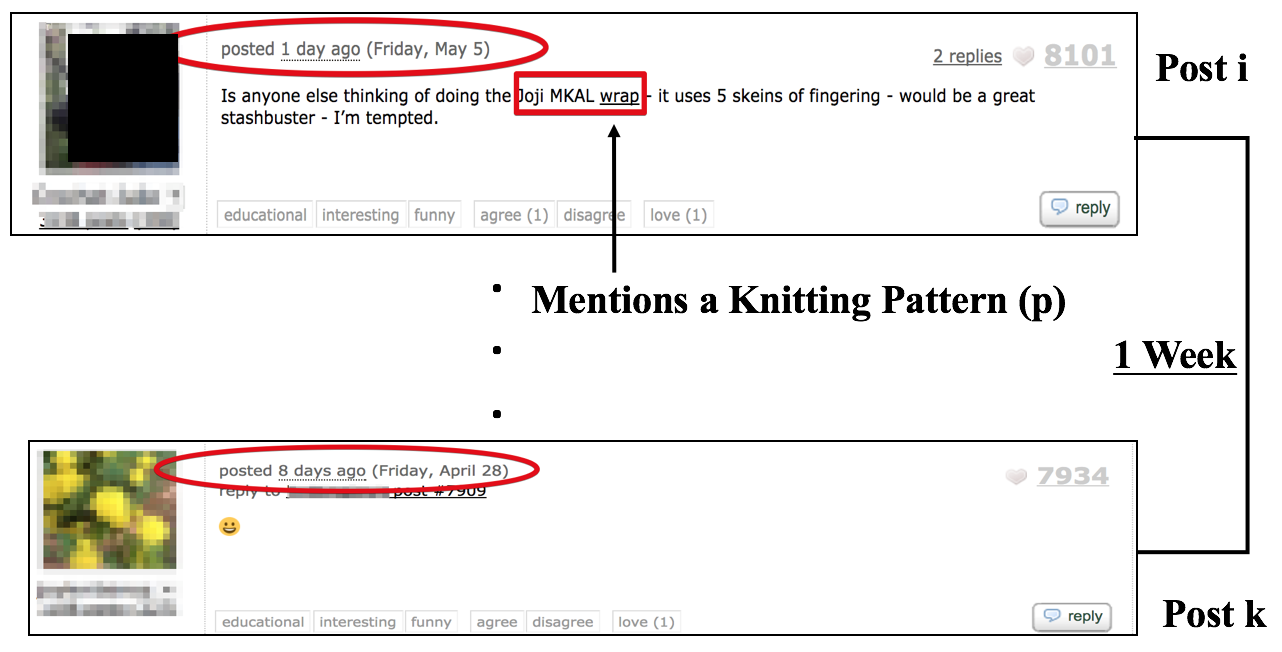}
    \caption{Operationalization of Exposure}
    \label{operation}
\end{figure}

Figure \ref{operation} shows how we operationalize exposure. The \textbf{exposure} of a post reflects the approximate number of users who read the post. There is no direct way to know who read a particular post. However, we have the information of the users who posted on the same thread and the time stamps of the posts. The traffic varies across different forums. By analyzing this traffic, we came up with the following heuristic to identify the number of people exposed to a post: we observed that people mention reading posts most frequently within a week of the post time. Posts older than a week cease to garner attention. Thus, we define exposure as:

\textit{If a post $\mathbf{i}$ mentioned a knitting pattern $\mathbf{p}$, then we consider all the users who posted to the thread up to one week after post $\mathbf{i}$ as exposed to post $\mathbf{i}$.}

\subsubsection{Uptake}
Another heuristic that we use to label the posts is \textbf{uptake}. For each user, we check if she/he used the pattern mentioned in the post or added it to her/his knitting queue after she/he was exposed to the post. Uptake reflects the percentage of exposed users who used the knitting pattern mentioned in the post or added it to their knitting queue\footnote{Users can maintain a knitting queue, where they add their future projects and information about the materials and the pattern they plan to use. They might use the pattern without adding it to their queue}. Uptake is defined as: 

\textit{Let $\mathbf{x}$ denote the number of users who were exposed to the post $\mathbf{i}$ and used the knitting pattern $\mathbf{p}$ mentioned in the post. Let the total number of users exposed to the post $\mathbf{i}$ be $\mathbf{n}$. Then for the post $\mathbf{i}$, percent uptake is $\mathbf{x}$/ $\mathbf{n}$ * 100}

Therefore, uptake is the percentage of users exposed to post $\mathbf{i}$ that took up a knitting pattern mentioned in it. In our experiments, if the percent uptake of post $\mathbf{i}$ is greater than 0, we label the post as \textit{influential} otherwise we label it as non-influential. With this approach, the raw data consisted of 34.10\% influential posts and 65.90\% of non-influential posts. A subset of this dataset was sampled for manual annotations for our experiments, described in detail in the following sections. A total of 700 posts were selected, with 340 influential and 360 non-influential posts.

\section{Annotation of Influence}
In order to identify and distinguish the linguistic characteristics of influential vs non-influential posts, we look for language features motivated from the basic principles of social influence. In a platform like Ravelry, the principle of `Social Validation' is an undercurrent of people's activities across different groups. \citeauthor{cialdini2002science} define social validation as a phenomenon in which people frequently look to others for cues on how to `think', `feel' and `behave'. In our experiments, we operationalize influence assuming that people take cues from influential posts in order to think and decide on which pattern to use. 

\textbf{Theoretical grounding of cues.} In order to model the presence of these influential cues, we must understand the novelty of the language used to present the pattern. In a post, excitement reflects the happiness experienced by the member while using a pattern. Consequently, this cue motivates another member to use that pattern. Similarly, a detailed description of a pattern by using enhancing qualifiers, makes a post more attractive and triggers `liking' towards that pattern. Using different materials (yarn or fiber) or creating a modified version of the original pattern reflects the interest and the effort that a user puts into a pattern. A display of creativity makes a pattern more attractive by looking `new' and `different' and in turn motivates others to adapt the modified pattern.

We qualitatively looked at 50 influential and 50 non-influential posts for language cues that can make a post interesting to users. Based on our analysis, we propose three features that act as markers of these cues. These features are: `Enthusiasm', `Qualifiers' and `Modification'. The following sub-sections analyze each of these, provide examples, and explain how they are motivated from the basic principles of influence.

\subsection{Enthusiasm}
Enthusiasm is defined as a person's excitement and its intensity as displayed in a post. In influential posts, the expressed emotion is strongly positive. We focus on enthusiasm that is expressed towards a knitting pattern, project, yarn or related entities. If a user seems excited about these entities, that might entice others to be interested in the object of enthusiasm, as accordance with the social validation principle. We ignore enthusiasm expressed towards other users and entities not connected to the knitting project. In order to quantify the intensity of excitement, we look for punctuation markers (specifically exclamation marks) qualifying the statement with positive valence. Some examples of enthusiastic and non-enthusiastic posts are:

\begin{itemize}
\item Enthusiastic
\begin{itemize}
\item \textit{Yours \textbf{look really great!} And that reminds me that I never posted mine in this group! :) So here they are.}
\item \textit{Cable mittens. Knit flat and seamed - \textbf{an easy way to make thumbs!} Of course you have to seam them, but you can barely see the seam on the moss stitch.}
\end{itemize}
\item Non-Enthusiastic
\begin{itemize}
\item \textit{I enjoyed making this Hue Shift afghan so much that I am sure I will make another. By the way, the camera picks the red up, in real life the red does not form a cross.} $\rightarrow$ intensity of excitement is \textbf{low}.
\item \textit{;-P sun came out today! The camera was set for flash and that was the best photo, so the cable work is very visible. The design did stand out more as more work was done, but it still doesn,t seem to pop as much as the other images here.} $\rightarrow$ excitement shown is \textbf{not for the pattern} or related entity
\end{itemize}
\end{itemize}

\subsection{Qualifiers}
Qualifiers are words or phrases that provide descriptive details that enhance the impact of the description of a pattern. Qualifiers can either highlight a pattern's quality or usability, features of the yarn or the stitches used, color effects and more. Some example phrases are: \textit{`quick and easy to follow'}, \textit{`perfect pattern'}, \textit{`super-soft handspun yarn'}. 

Therefore, qualifiers hint at the attractiveness and usability of the pattern or the yarn. A post that presents the pattern in a positive light with these qualifiers may exert influence to adapt them, consistent with the `liking’ principle. The following example posts illustrate valid qualifiers:


\begin{itemize}
\item \textit{This is the cuff of the left mitt, couldn’t stop and finished clue 2 before I took a picture of both cuffs. But I like how the Zauberball comes out, they wont be identical, but I love them. Thanks, Paula, love the pattern and how you wrote it. \textbf{It’s really easy going}.}
\item \textit{I’ve been asked to knit fingerless gloves for my 3 nephews Christmas gifts. I did the first pair using the 75 Yard Malabrigo Mitts in two yarns, I am half done with the 2nd pair in the same pattern and have yet to start the 3rd. \textbf{That pattern seems non-gender-specific}.}
\item \textit{This is \textbf{such a nice pattern}! I knitted them last september for a swap: And I probably knit another pair for me soon :-)} $\rightarrow$ \textit{(Both \textbf{enthusiastic} and has a \textbf{pattern qualifier})}
\item \textit{Two patterns that I know of that \textbf{handle highly variegated yarns} are Aquaphobia and Harvest Dew. This one also \textbf{looks interesting}, Indiana Jones and the Socks}.
\end{itemize}

\subsection{Modification}
Modification captures the actual or the suggested changes made to an original pattern. Some examples of the changes that modification attempts to capture are: 
\begin{itemize}
\item Adding or removing rows
\item Changing the size or the shape of the pattern
\item Using extra or lesser stitches
\item Using different needles for stitching
\item Adding or omitting something from the pattern
\item Processing the yarn in a particular manner
\end{itemize}
This set of descriptive modifiers does not include the number of days, the effort put in the completion of the pattern or the quantity of materials required. These are not included because they vary by user but do not offer much insight into the creativity of the user.

As mentioned before, the principle of `Social Validation' states that people often look to others in order to decide if and how to modify their behavior. Modifications exemplify an individual's creativity and interest in a pattern. By this principle, these described changes to the pattern might in turn influence other users to adopt the pattern. Some example posts that denote modification are:
\begin{itemize}
\item \textit{Pattern: Maize -Yarn is Cascade 220 Heathers, color 9452, 103 yards -Any modifications to the pattern: \textbf{one extra row on the thumb for length} -This was my first mitten and I see many more in my future! The pattern was very straightforward, as it is designed for beginners.}
\item \textit{These are \textbf{my version of} the oh-so-popular Fetching. I can see why they’re so popular: well-written pattern and clever use of cables. I \textbf{cast on 40 and did an extra cable repeat at the top}.hand model but should be comfy on the 11-year-old recipient}
\item \textit{Yours look great, but if I personally were doing them, I \textbf{would modify them} to look like this: It \textbf{should be possible} to keep the colorwork regular even with decreases. I would look at the Egyptian Mittens, etc.}
\item \textit{Pattern: Norwegian Selbu Mittens -Yarn: Dalegarn Heilo - 1 skein charcoal \& 3/4 skein red -Project page: Sochi Selbu -Mods: \textbf{Added some Xs for shorter floats and plain thumb tips} -Notes: Decided I prefer this stickier yarn for stranded knitting.}
\end{itemize}

\section{Experiments and Results}
Two annotators labeled 700 posts with the presence of enthusiasm, qualifier and modification cues. The actual class labels \textbf{(Influential or Non-Influential)} were not revealed to them for the annotation process. In order to evaluate the robustness of these annotations, we measured the inter-annotator agreement by computing Cohen's Kappa for a subset of 40 commonly annotated posts (different from the 700 posts mentioned above). We got satisfactory agreement between the annotators on the definition of our linguistic cues. The kappa values for the two annotators are shown in Table \ref{annotator}. \newline

\begin{table}[htbp]
\setlength\tabcolsep{3pt}
\centering
\renewcommand{\arraystretch}{1}
\begin{tabular}{|p{3cm}| l | r|}
\hline
Label & Cohen's Kappa \\
\hline
Enthusiasm &  0.7333 \\
Qualifiers & 0.9310 \\
Modification & 0.7561 \\
\hline
\end{tabular}
\caption{Inter Annotator Agreement Values}
\label{annotator}
\end{table}

We performed experiments to automatically classify the posts with their influence label using our features in a machine learning model. The classifier gives an insight into the predictive power and the robustness of the linguistic features described in Section 4. 

As discussed earlier, we classify the posts in two classes: `\textit{Influential}' and `\textit{Non-influential}'. The baseline model is a logistic regression classifier with L2 regularization that uses `\textbf{Unigram}' features only. The binary labels for `modification', `enthusiasm' and `qualifiers' \textbf{(MEQ)}, as identified by the annotators, are then included in addition to the unigram features. MEQ also includes four other features constructed by combining the individual binary features. In particular, this includes: `enthusiasm and qualifier', `enthusiasm and modification', `enthusiasm and qualifier and modification' and `qualifier and modification'. These combination features, or interaction terms, are important. For example, enthusiasm alone might not be sufficient to spark an interest in the user so as to influence her/him into adopting a knitting pattern. A post that emphasizes the qualities of a pattern or details the different variations possible for a pattern along with an undercurrent of enthusiasm, makes a pattern more attractive than the one with just an enthusiastic emotion.


\textbf{Word-Category based features:} \citet{persuasion}'s earlier work on persuasion used word categories (WC) as features for identifying persuasiveness in text. We explore similar categories like `pronoun counts', `raw number of word occurrences', `count of articles in the post', `length of the post' and more (See Table \ref{weight}) as features for our experiments. We used the python readability calculator to estimate these features.\footnote{https://pypi.python.org/pypi/readability/0.1}

\textbf{Sentiment based features:} As mentioned in Section 2, sentiment or the way people `feel' plays an important role in interpersonal interactions. Hence, we use sentiment features calculated by using a sentiment analyzer from \citet{hutto2014vader}. The tool estimates four scores for each post: `positive', `negative', `neutral' and `compound'. The positive, neutral and negative score represent the proportions of the text that fall into each of these categories respectively. The compound score aggregates the overall sentiment of the post.

In order to have a fair comparison, we used a logistic regression classifier with L2 regularization and 5 fold cross-validation for all our experiments, which were performed using Lightside \cite{mayfield2013lightside}. The results are shown in Table \ref{results}. The columns report the `Accuracy' and `Cohen's Kappa' values for different feature sets (Unigram, MEQ, WC and Sentiment). These experiments were performed in order to validate the contribution of our MEQ features for predicting social influence.

\begin{table}[htbp]
\setlength\tabcolsep{3pt}
\centering
\renewcommand{\arraystretch}{1}
\begin{tabular}{| p{4cm} | c | r |}
\hline
Model & Accuracy  & Kappa\\
\hline
Unigram &  68.71 & 0.3735 \\
\hline
Unigram + MEQ & 69.14 & 0.3825 \\
\hline
Unigram + Sentiment & 69.29 & 0.3850 \\
\hline
Unigram + WC & 71.43 & 0.4280 \\
\hline
Unigram + WC + Sentiment & 71.14 & 0.4223 \\
\hline
Unigram + WC + MEQ & 71.57 & 0.4304 \\
\hline
\textbf{Unigram + WC + Sentiment + MEQ }& \textbf{71.86} & \textbf{0.4361} \\
\hline
\end{tabular}
\caption{Accuracy and Kappa Results}
\label{results}
\end{table}

Accuracy may not be a sufficient metric to capture specifically what the model learned about the positive (Influential) class. It is possible that the accuracy is high because the model learned to predict the negative class (Non-Influential) correctly. In order to make this distinction, we look at the confusion matrix shown in Table \ref{confuse}. The table shows a comparison between the true positives of the baseline model and those of the best performing model along with the respective F-Scores. The true positives are the influential posts in our data labeled as defined in Section 3. The predicted positives are posts that were predicted as influential by the model. A similar definition stands for true negative and predicted negative. As shown in the table, the model trained on all the feature sets, correctly classifies more positive labels than the baseline model. Hence, we also get an improvement of 2.91 point for F-score.

\begin{table}[htbp]
\setlength\tabcolsep{3pt}
\centering
\renewcommand{\arraystretch}{1}
\begin{tabular}{| p{0.5cm} | p{0.6cm} | p{0.6cm} | p{1.0cm}| p{3.75cm} |}
\hline
 & PN  & PP & F & Model \\
\hline
TN & 253 & 107 & 67.55 & Unigram\\
TP & 112 & 228 & & \\
\hline
TN & 267 & 93 & 70.46 & Unigram + WC \\
TP & 103 & 237 & & + Sentiment + MEQ  \\
\hline
\end{tabular}
\caption{Confusion Matrix for baseline and the best performing model. In the table, TN=True Negative, TP=True Positive, PN=Predicted Negative, PP=Predicted Positive and F=F-Score.}
\label{confuse}
\end{table}

\section{Discussion}

The results presented above suggest that the added features play a role in achieving influence.  Here we offer more insight through posthoc analysis.  First we explore feature weights. Table \ref{weight} shows the feature weights for the identified significant features. 


As we can see, `Qualifier' gets a high feature weight. From our discussion in Section 4, we know that the posts with qualifier cues hint at the attractiveness and likability of the patterns by providing descriptive details about them. This suggests that the `Liking' principle, on which the `Qualifier' feature is based, plays a pivotal role in explaining influence in interpersonal interactions. However, `Enthusiasm' has a lower weight than other features. In fact, `Enthusiasm' alone might not be sufficient to predict the label of a post. However, the combinations of these features, specifically `Enthusiasm and Modification' has a particularly high weight. This implies that, if the author of the post makes some modifications to the pattern and seems enthusiastic about it, the users exposed to the post might have a higher chance of getting interested and adapting the pattern. The principle of `Social Validation' is therefore portrayed well by this interaction feature. The high weight is in line with our expectation that this principle is an important undercurrent of user activities on Ravelry.

We can observe from Table \ref{results} that the MEQ features improve the accuracy of the model. The feature weights shown in Table \ref{weight} suggests that some of these features, have high positive weights and some have higher weights than the word-category features, hinting that they might be better predictors of influence than the WC features. The word-category features capture the number of pronouns, nominalizations, articles, subordination and more. These elements are not covered by any of our MEQ features. 

\begin{table}[htbp]
\setlength\tabcolsep{3pt}
\centering
\renewcommand{\arraystretch}{1}
\begin{tabular}{ l | r }
\hline
Feature Name & Weight\\
\hline
\textbf{MEQ and derived features} & \\
Qualifier & 0.8277 \\
Enthusiasm and Modification & 0.7907 \\
Modification & 0.3998 \\
Enthusiasm and Qualifier & \\
and Modification & 0.3557 \\
Qualifier and Modification & 0.1082 \\
Enthusiasm and Qualifier & 0.0037 \\
Enthusiasm & -0.1946 \\
& \\
\textbf{Word category-based features} & \\
tobeverb & 0.4921\\
nominalization & 0.2951\\
complex\_wrds\_dc & 0.1082\\
post length & 0.0237\\
article & -0.2482\\
subordination & -0.2746\\
pronoun & -1.1194\\
& \\
\textbf{Sentiment-based features} & \\
compound & 0.2886 \\
positive & 0.0912 \\
neutral & 0.0749 \\
negative & -0.2121 \\
\hline
\end{tabular}
\caption{Feature weights for important features}
\label{weight}
\end{table}

\textbf{Forward Feature Selection:} In any model with a large variety and number of low level features, there may be many correlated features that share weight, and thus we cannot properly interpret the observed weights.  One way of isolating the value of specific features is to do a forward feature selection and identify which features are selected for the optimal set. We ran a series of such experiments, varying the number of features to select from \textbf{900 to 200}.  In all cases, the four interaction terms for our MEQ features (`enthusiasm and modification', `qualifier and modification', `enthusiasm and qualifier' and `enthusiasm and qualifier and modification')
along with the individual feature `Qualifier' were selected as prominent predictors of influence.  Even with the smallest resulting feature set, the classification accuracy remained at \textbf{71\% }.  This supports the value placed on our added features by the weight analysis above.

\textbf{Error Analysis: }In order to understand the limitations of the MEQ features, we performed error analysis on our model. The following example shows an influential post that was wrongly predicted as non-influential by the model:
\newline
``\textit{This \textbf{KAL is coming at the right time wonderful!} Need to finish some WIPs: \underline{kalajoki} which shall become a christmas gift puzzle socks - one down, one to go kleinkariert I and kleinkariert II. I would be glad to join you}.''

The enthusiasm displayed in the post is not towards the pattern (\underline{\textit{kalajoki}}) itself. The post is enthusiastic about a KAL, which is a `Knit Along' event occurring in the group. The users might have a greater tendency to adapt patterns during KAL and similar events. In cases like this, the measured influence of a post might be affected by other contextual factors like the occurrence of a KAL. In order to incorporate these behaviors in the classification model, a better understanding of the group dynamics is required. We leave this to subsequent work.

Following is another example of an influential post predicted as non-influential by the model: \newline
``\textit{This is what I choose, what do you think? In the second \underline{\textbf{picture}} I put some other shades of yellow/orange; green; grey/blue}'' 

Even though the post is marked as influential in the data, the language of the post does not contain cues for either enthusiasm or qualifiers or modification. The attractiveness of the pattern might have been captured in the picture in the post and not in the text itself. Such noise exists in our data.

\textbf{The Homophily Confound:} \citet{shalizi2011homophily} identify three factors that affect the activities in a social network: `Homophily', `Social Influence' and `Co-Variate Causation'. It is difficult to distinguish between them. Homophily occurs when social ties are formed among people due to similar individual traits and choices. It is difficult to identify if two people chose the same pattern because they like similar things (homophily) or because one influenced the other (social influence). We have not addressed this problem in the current setup and hope to explore it in the future.

\section{Conclusion and Future Work}

In this paper, we have studied social influence in an online community setting featuring interpersonal interactions. 
We designed an approach to operationalize influence in this setting and a task that enables us to measure the impact of textual features on influence. We presented three new features that are motivated from theoretical principles found in the literature on social influence. Adding them to a baseline model, we achieved an improvement of 3.15\% in accuracy and  2.91 points in F-score with our final F-score being 70.46\%.

In the future, we would like to further study influence in interpersonal interactions along three directions. Firstly, we would like to study influence in interpersonal interactions of groups that have different goals and interests. Secondly, we would like to study the ways in which the other principles of influence come into play for interpersonal interactions. This study focused on the principles of `Social Validation' and `Liking'. The remaining principles might give a different view of influence among people. For example, the principle of `authority' might come into play when a moderator or an experienced person in a group recommends a pattern. Similarly, there might be an influence among people due to `reciprocation' depending on the history of their activities in different groups. It would be interesting to explore such principles through the various activities on the Ravelry platform. Thirdly, as discussed earlier, we would like to tease apart the effects of `Homophily' and `Social Influence' while studying the spread of pattern usage in Ravelry.

\section{Acknowledgement}
This work was funded in part by NSF grant IIS 1546393, a fellowship from Bosch and DARPA grant FA8750-12-2-0342.

\bibliographystyle{acl_natbib}

\begin{thebibliography}{}
\expandafter\ifx\csname natexlab\endcsname\relax\def\natexlab#1{#1}\fi

\bibitem[{Anagnostopoulos et~al.(2008)Anagnostopoulos, Kumar, and
  Mahdian}]{anagnostopoulos2008influence}
Aris Anagnostopoulos, Ravi Kumar, and Mohammad Mahdian. 2008.
\newblock Influence and correlation in social networks.
\newblock In {\em Proceedings of the 14th ACM SIGKDD international conference
  on Knowledge discovery and data mining\/}. ACM, pages 7--15.

\bibitem[{Anger and Kittl(2011)}]{anger2011measuring}
Isabel Anger and Christian Kittl. 2011.
\newblock Measuring influence on twitter.
\newblock In {\em Proceedings of the 11th International Conference on Knowledge
  Management and Knowledge Technologies\/}. ACM, page~31.

\bibitem[{Bakshy et~al.(2011)Bakshy, Hofman, Mason, and
  Watts}]{bakshy2011everyone}
Eytan Bakshy, Jake~M Hofman, Winter~A Mason, and Duncan~J Watts. 2011.
\newblock Everyone's an influencer: quantifying influence on twitter.
\newblock In {\em Proceedings of the fourth ACM international conference on Web
  search and data mining\/}. ACM, pages 65--74.

\bibitem[{Bales(1956)}]{bales1956task}
Robert~F Bales. 1956.
\newblock Task status and likeability as a function of talking and listening in
  decision-making groups.
\newblock {\em The state of the social sciences\/} pages 148--161.

\bibitem[{Bales(1973)}]{bales197315}
Robert~F Bales. 1973.
\newblock 15 robert f. bales the equilibrium problems in small groups.
\newblock {\em Social encounters: readings in social interaction\/} page 221.

\bibitem[{Biran et~al.(2012)Biran, Rosenthal, Andreas, McKeown, and
  Rambow}]{biran2012detecting}
Or~Biran, Sara Rosenthal, Jacob Andreas, Kathleen McKeown, and Owen Rambow.
  2012.
\newblock Detecting influencers in written online conversations.
\newblock In {\em Proceedings of the Second Workshop on Language in Social
  Media\/}. Association for Computational Linguistics, pages 37--45.

\bibitem[{Cha et~al.(2010)Cha, Haddadi, Benevenuto, and
  Gummadi}]{cha2010measuring}
Meeyoung Cha, Hamed Haddadi, Fabricio Benevenuto, and P~Krishna Gummadi. 2010.
\newblock Measuring user influence in twitter: The million follower fallacy.
\newblock {\em Icwsm\/} 10(10-17):30.

\bibitem[{Chen et~al.(2009)Chen, Wang, and Yang}]{chen2009efficient}
Wei Chen, Yajun Wang, and Siyu Yang. 2009.
\newblock Efficient influence maximization in social networks.
\newblock In {\em Proceedings of the 15th ACM SIGKDD international conference
  on Knowledge discovery and data mining\/}. ACM, pages 199--208.

\bibitem[{Cialdini and Goldstein(2002)}]{cialdini2002science}
Robert~B Cialdini and Noah~J Goldstein. 2002.
\newblock The science and practice of persuasion.
\newblock {\em The Cornell Hotel and Restaurant Administration Quarterly\/}
  43(2):40--50.

\bibitem[{Cialdini and Goldstein(2004)}]{cialdini2004social}
Robert~B Cialdini and Noah~J Goldstein. 2004.
\newblock Social influence: Compliance and conformity.
\newblock {\em Annu. Rev. Psychol.\/} 55:591--621.

\bibitem[{Goyal et~al.(2010)Goyal, Bonchi, and Lakshmanan}]{goyal2010learning}
Amit Goyal, Francesco Bonchi, and Laks~VS Lakshmanan. 2010.
\newblock Learning influence probabilities in social networks.
\newblock In {\em Proceedings of the third ACM international conference on Web
  search and data mining\/}. ACM, pages 241--250.

\bibitem[{Goyal et~al.(2011)Goyal, Bonchi, and Lakshmanan}]{goyal2011data}
Amit Goyal, Francesco Bonchi, and Laks~VS Lakshmanan. 2011.
\newblock A data-based approach to social influence maximization.
\newblock {\em Proceedings of the VLDB Endowment\/} 5(1):73--84.

\bibitem[{Hutto and Gilbert(2014)}]{hutto2014vader}
Clayton~J Hutto and Eric Gilbert. 2014.
\newblock Vader: A parsimonious rule-based model for sentiment analysis of
  social media text.
\newblock In {\em Eighth international AAAI conference on weblogs and social
  media\/}.

\bibitem[{Katz and Lazarsfeld(1966)}]{katz1966personal}
Elihu Katz and Paul~Felix Lazarsfeld. 1966.
\newblock {\em Personal Influence, The part played by people in the flow of
  mass communications\/}.
\newblock Transaction Publishers.

\bibitem[{Mayfield and Ros{\'e}(2013)}]{mayfield2013lightside}
E~Mayfield and CP~Ros{\'e}. 2013.
\newblock Lightside: Open source machine learning for text accessible to
  non-experts. invited chapter in the handbook of automated essay grading.

\bibitem[{Ng et~al.(1993)Ng, Bell, and Brooke}]{ng1993gaining}
Sik~Hung Ng, Dean Bell, and Mark Brooke. 1993.
\newblock Gaining turns and achieving high influence ranking in small
  conversational groups.
\newblock {\em British Journal of Social Psychology\/} 32(3):265--275.

\bibitem[{Ng et~al.(1995)Ng, Brooke, and Dunne}]{ng1995interruption}
Sik~Hung Ng, Mark Brooke, and Michael Dunne. 1995.
\newblock Interruption and influence in discussion groups.
\newblock {\em Journal of Language and Social Psychology\/} 14(4):369--381.

\bibitem[{Nguyen et~al.(2014)Nguyen, Boyd-Graber, Resnik, Cai, Midberry, and
  Wang}]{nguyen2014modeling}
Viet-An Nguyen, Jordan Boyd-Graber, Philip Resnik, Deborah~A Cai, Jennifer~E
  Midberry, and Yuanxin Wang. 2014.
\newblock Modeling topic control to detect influence in conversations using
  nonparametric topic models.
\newblock {\em Machine Learning\/} 95(3):381--421.

\bibitem[{Quercia et~al.(2011)Quercia, Ellis, Capra, and
  Crowcroft}]{quercia2011mood}
Daniele Quercia, Jonathan Ellis, Licia Capra, and Jon Crowcroft. 2011.
\newblock In the mood for being influential on twitter.
\newblock In {\em Privacy, Security, Risk and Trust (PASSAT) and 2011 IEEE
  Third Inernational Conference on Social Computing (SocialCom), 2011 IEEE
  Third International Conference on\/}. IEEE, pages 307--314.

\bibitem[{Reid and Ng(2000)}]{reid2000conversation}
Scott~A Reid and Sik~Hung Ng. 2000.
\newblock Conversation as a resource for influence: Evidence for prototypical
  arguments and social identification processes.
\newblock {\em European Journal of Social Psychology\/} 30(1):83--100.

\bibitem[{Rogers(2010)}]{rogers2010diffusion}
Everett~M Rogers. 2010.
\newblock {\em Diffusion of innovations\/}.
\newblock Simon and Schuster.

\bibitem[{Rosenthal and McKeown(2016)}]{rosenthal2016social}
Sara Rosenthal and Kathleen McKeown. 2016.
\newblock Social proof: The impact of author traits on influence detection.
\newblock In {\em Proceedings of the 1st Workshop on Natural Language
  Processing and Computational Social Science\/}. pages 27--36.

\bibitem[{Shalizi and Thomas(2011)}]{shalizi2011homophily}
Cosma~Rohilla Shalizi and Andrew~C Thomas. 2011.
\newblock Homophily and contagion are generically confounded in observational
  social network studies.
\newblock {\em Sociological methods \& research\/} 40(2):211--239.

\bibitem[{Shuai et~al.(2012)Shuai, Ding, Busemeyer, Chen, Sun, and
  Tang}]{shuai2012modeling}
Xin Shuai, Ying Ding, Jerome Busemeyer, Shanshan Chen, Yuyin Sun, and Jie Tang.
  2012.
\newblock Modeling indirect influence on twitter.
\newblock {\em International Journal on Semantic Web and Information Systems
  (IJSWIS)\/} 8(4):20--36.

\bibitem[{Sim et~al.(2016)Sim, Routledge, and Smith}]{simfriends}
Yanchuan Sim, Bryan~R Routledge, and Noah~A Smith. 2016.
\newblock Friends with motives: Using text to infer influence on scotus .

\bibitem[{Tan et~al.(2016{\natexlab{a}})Tan, Niculae, Danescu-Niculescu-Mizil,
  and Lee}]{tan2016winning}
Chenhao Tan, Vlad Niculae, Cristian Danescu-Niculescu-Mizil, and Lillian Lee.
  2016{\natexlab{a}}.
\newblock Winning arguments: Interaction dynamics and persuasion strategies in
  good-faith online discussions.
\newblock In {\em Proceedings of the 25th International Conference on World
  Wide Web\/}. International World Wide Web Conferences Steering Committee,
  pages 613--624.

\bibitem[{Tan et~al.(2016{\natexlab{b}})Tan, Niculae, Danescu-Niculescu-Mizil,
  and Lee}]{persuasion}
Chenhao Tan, Vlad Niculae, Cristian Danescu-Niculescu-Mizil, and Lillian Lee.
  2016{\natexlab{b}}.
\newblock Winning arguments: Interaction dynamics and persuasion strategies in
  good-faith online discussions.
\newblock In {\em Proceedings of WWW\/}.

\bibitem[{Tang et~al.(2009)Tang, Sun, Wang, and Yang}]{tang2009social}
Jie Tang, Jimeng Sun, Chi Wang, and Zi~Yang. 2009.
\newblock Social influence analysis in large-scale networks.
\newblock In {\em Proceedings of the 15th ACM SIGKDD international conference
  on Knowledge discovery and data mining\/}. ACM, pages 807--816.

\bibitem[{Weng et~al.(2010)Weng, Lim, Jiang, and He}]{weng2010twitterrank}
Jianshu Weng, Ee-Peng Lim, Jing Jiang, and Qi~He. 2010.
\newblock Twitterrank: finding topic-sensitive influential twitterers.
\newblock In {\em Proceedings of the third ACM international conference on Web
  search and data mining\/}. ACM, pages 261--270.

\end{thebibliography}

\end{document}